%% file: acl_latex.tex
\pdfoutput=1

\documentclass[11pt]{article}

\usepackage[final]{acl}

\usepackage{times}
\usepackage{latexsym}
\usepackage{xspace}
\usepackage{makecell}

\usepackage[most]{tcolorbox}
\usepackage[T1]{fontenc}

\definecolor{beaublue}{rgb}{0.74, 0.83, 0.9}

\usepackage[utf8]{inputenc}
\usepackage{graphicx}
\usepackage{subcaption}

\usepackage{microtype}

\usepackage{inconsolata}

\usepackage{microtype}

\usepackage{inconsolata}
\usepackage{tcolorbox}
\usepackage{booktabs}       
\usepackage{amsfonts}       
\usepackage{nicefrac}       
\usepackage{microtype}      
\usepackage{xcolor}         
\usepackage{xspace}
\usepackage{amsmath}
\usepackage{optidef}
\usepackage{comment} 
\usepackage{multirow}
\usepackage{amsthm}
\usepackage[normalem]{ulem}
\usepackage{wrapfig}

\usepackage{colortbl}
\usepackage{arydshln}

\newcommand{\aspace}{\hspace{1em}}
\newcommand{\uw}{$^{1}$}
\newcommand{\copenhagen}{$^{2}$}
\newcommand{\chipstack}{$^{3}$}
\newcommand{\stanford}{$^{4}$}
\usepackage[first=0,last=9]{lcg}

\definecolor{maria-color}{HTML}{7881F2}

\usepackage{graphicx}
\usepackage{tablefootnote}

\usepackage{xcolor}
\definecolor{gpt3-color}{HTML}{d4fae6}
\definecolor{gpt4-color}{HTML}{cff5f5}
\definecolor{llama-color}{HTML}{f9d4fa}

\title{Information-Guided Identification of Training Data Imprint in (Proprietary) Large Language Models}

\author{
    Abhilasha Ravichander \uw  \aspace 
    Jillian Fisher\uw  \aspace 
    Taylor Sorensen \uw  \aspace
    Ximing Lu \uw  \aspace \aspace\\
    \textbf{Yuchen Lin} \uw  \aspace  
    \textbf{Maria Antoniak} \copenhagen \aspace 
    \textbf{Niloofar Mireshghallah} \uw \aspace \\ 
    \textbf{Chandra Bhagavatula} \chipstack \aspace 
    \textbf{Yejin Choi}  \stanford \aspace \\
    \uw University of Washington \aspace
    \copenhagen University of Copenhagen \aspace 
     \chipstack ChipStack AI \aspace\\
    \stanford Stanford University\\[3pt]
    \texttt{aravicha@cs.washington.edu,yejinc@stanford.edu}
}
\begin{document}
\maketitle
\begin{abstract}
High-quality training data has proven crucial for developing performant large language models (LLMs). However, commercial LLM providers disclose few, if any, details about the data used for training. 
This lack of transparency creates multiple challenges: it limits external oversight and inspection of LLMs for issues such as copyright infringement, it undermines the agency of data authors, and it hinders scientific research on critical issues such as data contamination and data selection. How can we recover what training data is known to LLMs? In this work we demonstrate a new method to identify training data known to proprietary LLMs like \textsc{GPT}-4 without requiring any access to model weights or token probabilities, by using information-guided probes. Our work builds on a key observation: text passages with high surprisal are good search material for memorization probes.
By evaluating a model's ability to successfully reconstruct high-surprisal tokens in text, we can identify a surprising number of texts memorized by LLMs.\footnote{Code/data available at \url{https://github.com/AbhilashaRavichander/information-probing}}
\end{abstract}


\input{latex/sections/introduction.tex}

\input{latex/sections/methodology}
\input{latex/sections/experiments}
\input{latex/sections/experiments_fiction}
\input{latex/sections/experiments_nyt}

\input{latex/sections/experiments_contamination}

\input{latex/sections/analysis}

\input{latex/sections/discussion}

\input{latex/sections/related_work}

\section{Conclusion}
In the current landscape of closed LLMs, the lack of documentation surrounding training data remains a major obstacle for model auditing and for scientific exploration. In this work, we consider one of the most restrictive (and yet common) access scenarios: models which do not permit accessing pretraining data, model weights, or logits. We construct a probing strategy that only requires input-output access to a model, and that can be applied using much smaller and cheaper language models. We show that we can use these probes to identify documents in the training data of commercial LLMs. We hope this effort leads to greater transparency in the LLM ecosystem, and empowers data contributors to have greater agency when interacting with AI systems.

\section{Limitations}
We aim to highlight potential limitations of our work. First, we recognize that the task of training data identification can be framed in various ways. For instance, should paraphrases be treated as members or non-members of the training set? Similarly, what about derivative works of copyrighted content that may still retain information from the original source? How much of the content needs to be memorized to qualify as a member? As such, due to the inherent ambiguity in defining this task, we are uncertain about the precise ``nearness'' of the data we identify as memorized.

Second, our approach is heavily dependent on model memorization. As a result, if a model does not memorize the training data, our method will not be effective. Consequently, as we discuss in Section 5, extracting memorized data using reconstruction probing is likely to only be effective for large and capable instruction-tuned models.  We encourage future research to adapt our method to probes that do not rely on memorization, potentially by calibrating recovery rates using rare or generic tokens. Additionally, we do not leverage metadata about the text samples for any of the methods described in this work (such as authors of passages of fictional text). It is possible that leveraging such metadata can improve the precision of identifying memorized data. A further limitation is that we used datasets from previous studies, which introduces potential differences between members and non-members, such as temporal gaps or varying frequencies in the training data~\cite{duan2024membership}. We also acknowledge a limitation in the reference models used in our experiments: these models may not possess the same knowledge as the model being tested, which could introduce bias into results. In addition, these models may have already been trained on the sample being probed, which could affect the distribution of tokens that are identified as high-surprisal. An additional constraint of this method is the reliance on identifying high-surprisal tokens. It may be the case that a sample of text is sufficiently generic that such high-surprisal tokens cannot be found, or that spurious examples of high-surprisal tokens are found--- possibly due to limitations of the reference model. Here, a practitioner may want to explore alternative methods to identify memorized data.

Lastly, we recognize that closed-source models inherently exhibit some degree of variance, which can make it difficult to replicate our findings across different models or systems. Companies can also implement post-training strategies to circumvent the effectiveness of our method. As a result, applying our probes to future models may prove challenging, especially in environments where proprietary changes to models are not disclosed.

\section*{Acknowledgment}
The authors would like to thank Aakanksha Naik and Yanai Elazar for helpful discussions regarding this work. This research was supported by the NSF DMS-2134012, ONR N00014-24-1-2207, and the Allen Institute for AI.

\bibliography{custom}

\appendix
\input{latex/sections/appendix}


\end{document}

%% file: latex/sections/introduction.tex
\section{Introduction}

\begin{figure}[t]
\includegraphics[width=8cm]{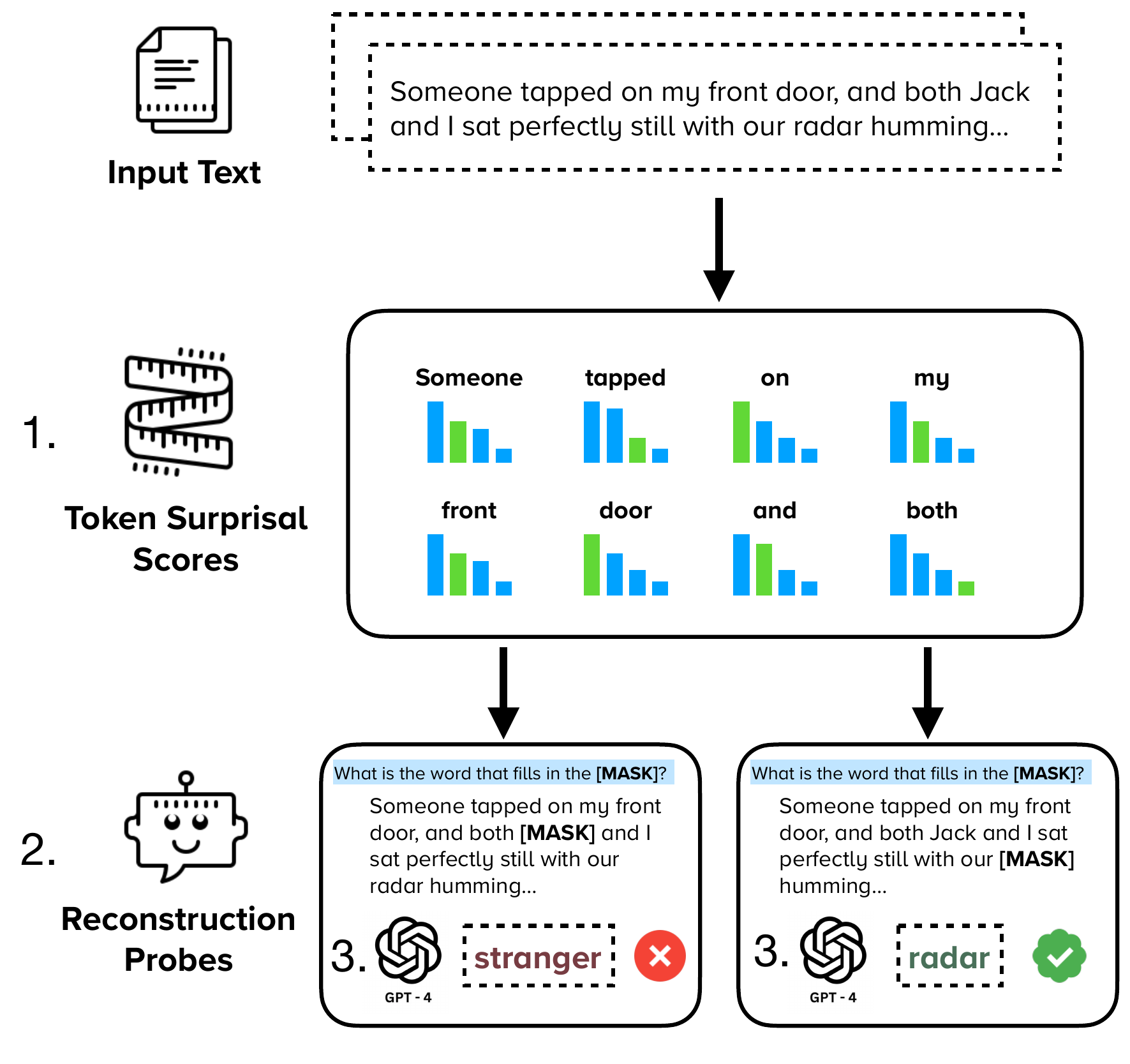}
\caption{Information-guided probes to identify training data. The probing  pipeline involves (1) finding \emph{surprising} tokens (tokens which are difficult to predict based on context), which can be accomplished using multiple approaches including leveraging domain knowledge, or relying on an external reference model, (2) constructing reconstruction probes where high-surprisal tokens are masked out and surrounding context tokens are kept constant, and (3) measuring the reconstruction rate for a given target model, i.e., the number of successful reconstructions of masked tokens. 
}
\end{figure}

For proprietary, legal, and reputational reasons, it has become common practice for companies to release few, if any, details about the secret ingredient --- training data --- powering their large language models (LLMs).
For example, the data used to train Gemini is described only at a high level as containing ``data from web documents, books, and code'' \citep{team2023gemini}, while Llama-2 apparently uses a ``new mix of data from publicly available sources, which does not include data
from Meta’s products or services'' \citep{touvron2023llama2}.
And even though training data is widely regarded as one of the most valuable components in building high-performing LLMs~\cite{team2023gemini}, most companies only provide access to the overarching model without detailed information about data sources or distributions, and certainly do not provide direct access to the model training data.




This lack of transparency has both scientific and societal implications, making it very difficult for 
(1)~researchers to evaluate the true capabilities and limitations of model generalization~\cite{balloccu-etal-2024-leak,aiyappa2024trustChatGPT}, 
(2)~external inspectors to examine models for data misuse~\cite{buick2024copyright,longpre2024data}, and
(3)~data authors to understand and control how their data is used~\cite{bommasani2023foundation,solove2024artificial}. 
To get around this issue, training data auditors have had to rely on roundabout and genre-specific memorization tests; for example, research investigating the use of published books as training data have used character name cloze tests \citep{chang2023speak}, which mask out a specific word in a text, or prefix completion methods to see if models will exactly generate a piece of text given the first few words of the text as a prefix~\cite{dsouza2023chatbot,karamolegkou-etal-2023-copyright,nytlawsuit}. However, cloze tests make assumptions about the pretraining data (e.g., the presence of character names) and prefix completion relies on heuristics for comparing model responses to the original text, as models do not always produce the completions verbatim.

In this work, we study the following problem: \emph{given a text sample and a closed, black-box model, is it possible to infer whether the sample may have been memorized by the model}? 
In this setting, a practitioner only has access to inputs and outputs from a model but no information about weights or logits; this is currently the most practical setting, as it mirrors the level of access provided to most commercial LLMs today. Crucially, unlike prior work, we focus not on specific data domains --- e.g., books~\cite{chang2023speak} or poems~\cite{walsh2024sonnet} --- but on datasets spanning texts from multiple domains. 
We develop tests that are applicable to diverse types of data, and that are more precise than simple prefix probing. 



To identify data that is known to models, without access to token probabilities, we construct probes based on the concept of \emph{surprisal}~\cite{shannon1948mathematical}: that a text passage may contain low-likelihood tokens that are difficult to reconstruct based on context alone. We then design cloze-style probes where these tokens are masked, and quantify the model's ability to reconstruct the tokens. Successful reconstruction would indicate that a model relies on one of two mechanisms: (1) reconstruction based on context, or (2) reconstruction based on memorization. If the token is chosen such that reconstruction based on context is challenging (for example, a minor character's name in a passage without any other identifying information), then the remaining mechanism for successful reconstruction is memorization.

These probes allow us to identify strong evidence of \emph{memorization}, finding that LLMs may be memorizing more, and different, data than what can be extracted by looking at a prefix completions alone. This is important because past work has shown that models may perform much better on memorized examples than they would otherwise~\cite{chang2023speak}, and there have been concerns that this problem may be affecting how LLMs are evaluated~\cite{oren2023proving}. Our work sheds light on the training data imprints in LLMs, and the potential risks of this memorization such as test set contamination. We hope our work fosters a culture of greater data transparency in the LLM ecosystem.

%% file: latex/sections/methodology.tex
\section{Background}


\paragraph{Memorization tests.}
Memorization tests can be developed using two main assumptions; the method has access to underlying token probability distributions or it has no access the these distributions. In this paper, we focus on the latter as it is the most common setting seen with current LLMs. For this setting, we focus on \emph{discoverable memorization}, where given part of a training data sample as input, the model can recover the remainder of the sample~\cite{carlini2022quantifying,liu2023llm360,chang2023speak,dsouza2023chatbot}. Past work has shown that about 1\% of training datasets can be recovered in this way~\cite{carlini2022quantifying}. 
However, these methods have limitations; cloze tests have required that the data contain character names or other specific features \citep{chang2023speak}, and prefix completion tests are less effective. Modern large language models likely incorporate additional posttraining or output filters to safeguard against verbatim regurgitation, and hence prefix probing methods often rely on heuristics (such as matching the longest common subsequence between the model generation and the orginal text) to compare model completions~\cite{karamolegkou-etal-2023-copyright}.

\paragraph{Surprisal.}
\emph{Surprisal}~\cite{shannon1948mathematical}, or the information conveyed by a linguistic unit in context, is a widely used measure in the computational modeling of human language processing~\cite{hale-2001-probabilistic,levy2008expectation}. Briefly, the information content of a word $w_{t} \in V$ that occurs in its context $w_{<t} \in V$ can be denoted as:
\begin{align}
\text{I}(w_{t}) = -\log P(w_{t}|w_{<t})
\end{align}
Predictable linguistic units carry lower information and have lower surprisal, whereas units that are \emph{unexpected} transmit higher information, and thus have higher surprisal. Recently, LLMs have been studied as estimators for token-level surprisal, by examining their correlation with different psychometric variables that predict human comprehension behavior~\cite{mcdonald2003low,goodkind2018predictive,oh2023does,giulianelli-etal-2023-information}.
In this work, we use LLMs to identify tokens carrying high information, to separate two distinct mechanisms that models can use to reconstruct training data: (a) reconstructing from surrounding context of a token, or (b) reconstructing through memorization of training data.

\section{Methodology}

Our work aims to surface evidence of text samples that have been memorized by models, and we show that we can recover examples at higher precision compared to previous approaches such as prefix probing. Information-guided probes offer a complementary view of identifying training data that is memorized by models by identifying tokens that are challenging to predict without memorization. Concretely, information-guided probing involves first identifying high-surprisal tokens, then constructing probes, and finally quantifying the reconstruction capacity of target models. We discuss details of this probing pipeline.

\paragraph{Probe design.}
Taking inspiration from prior work \citep{chang2023speak}, we formulate our probes as a cloze task, where a single high-surprisal token is masked out, and a target model is prompted to predict the token that fills in the mask, as shown in Figure 1. The exact prompts to the model, including instructions (and in-context examples), are described in Appendix B. This allows our probes to not require access to token probabilities assigned by a model, instead only leveraging model generations from the target model. Tokens are masked out one at a time, while the remaining tokens in context are held constant. Finally, for a given target model, we examine the reconstruction rate, or the number of high-surprisal tokens the target model can reconstruct. 

\paragraph{Information measures for token selection.}
In practice, there are a variety of ways to measure high-information carrying tokens. We consider two different methods to measure the amount of information being transmitted by a token $x$.

(1) The probability of $w_{t}$ given context $c$, where $h_{c}$ is the hidden state of a model for the context $c$:
\begin{align}
\text{Prob}(w_{t}) = -\log P(w_{t}|h_{c})
\end{align}

(2) The rank of $w_{t}$ in a vocabulary space $V$ given context $c$, or the number of more plausible alternatives:
\begin{equation}
    \text{Rank}(w_{t}) = \left|\{x : P(x|h_{c}) > P(w_{t}|h_{c}), x \in V\}\right|
\end{equation}

The rank metric evaluates a token's position in the sorted probability distribution of all possible tokens at a given masked position, rather than just its raw probability. This captures information about how many other tokens could plausibly appear in that position. For example, consider a probability distribution where an alternative token $x$ is assigned the highest probability, receiving the vast majority of the probability mass, and the token $w_{t}$ is assigned the second highest probability. Here, while traditional probability-based surprisal would assign $w_{t}$ a high surprisal value due to its low likelihood, the rank metric would assign it a lower surprisal value since there is only one more contextually appropriate choice. Yet another way of identifying high-information tokens is by leveraging domain knowledge (as we see in \S4.1, this is the case for character names in fictional text). 

How do we extract these information measures? Practically, we use a secondary model, known as a \emph{reference model} to extract (2) and (3), since we do not have access to token probabilities from the target model.  In order to get accurate information measures, in an ideal scenario we would want this reference model to be one that has not memorized the datapoint in question, in order for the token probabilities to be correctly-calibrated and to not be influenced by training on that particular datapoint. In practice, it is infeasible to find a large language model that has not been trained for every data sample. Instead, we seek to use a low-capacity model as they memorize training data at significantly lower rates~\cite{carlini2022quantifying} to extract information measures. 

\paragraph{Accounting for context using knowledge filters.}
In real-world domains, identifying what is surprising often requires world knowledge, but different models may encode different sets of facts.  Thus, we employ an ensemble of models to correctly contextualize surprising knowledge. In practice, after obtaining candidate surprise tokens from a reference model, we filter out those tokens which can be correctly surmised by a secondary (or even an ensemble of secondary) low-capacity instruction-tuned LLM(s). We note that the knowledge filters are low-capacity instruction tuned models, which differ from the the reference model which is a base model. The instructions used for knowledge filters are described in Appendix B. We use the token probabilities from the reference models to identify candidates for surprisal tokens, whereas we prompt the knowledge filtering models to reconstruct these candidate tokens after masking them, in order to filter out which ones can be easily guessed in a setting that closely matches the final target model.


\paragraph{Putting it all together.} The full pipeline that combines these components is shown in Figure~1. To determine whether a piece of text has been memorized, we first identify high-surprisal tokens using a reference model. Optionally, a knowledge filter may be applied to filter spurious high-surprisal tokens. These tokens are then masked one at a time, and the target model is prompted to reconstruct the masked token. We then measure the reconstruction rate, or the number of successful reconstructions, to identify if text has been memorized by a model. 


%% file: latex/sections/experiments.tex
\section{Experiments}
We are interested in identifying whether some text $d$ has been memorized by a model $M$.  We investigate two distinct risks associated with model memorization: (1) the memorization of copyrighted content such as works of fiction or news articles (\textbf{Fiction} and \textbf{New York Times}), and (2) the contamination of evaluation metrics through direct memorization of test samples, which undermines the assessment of a model's capabilities by allowing it to succeed through recall rather than by applying intended skills (\textbf{Dataset Contamination}). 

We examine the performance of two closed models: GPT-3.5 (\texttt{gpt-3.5-turbo-0125}), GPT-4 (\texttt{gpt-4-0613}), and an open-weight model: Llama-2-70B~\cite{touvron2023llama2}. Note that as of the writing of this paper,  Llama-2-70B is open-weight, but details of its training data remain unknown. For all the described experiments, we use BERT (110M parameters) as the reference model. 
We hold out 1870 instances from BookMIA~\cite{shi2023detecting} to tune hyperparameters such as the thresholds for selecting surprising tokens. 
We select low-probability or highly ranked tokens for our two information measures (please refer to Appendix A for details of all hyperparameters). To evaluate detection methods, we primarily use precision (the proportion of correctly identified memorized samples among all samples flagged as memorized). When comparing methods with similar precision scores, we prefer those that identify more memorized samples. Therefore, we report the $F_{\beta=0.1}$ score which weights precision more highly than recall. 

For tasks which examine the memorization of copyrighted content (\textbf{Fiction} and \textbf{New York Times}), we compare against prefix probing due to it's widespread use~\cite{karamolegkou2023copyright,nytlawsuit}. Prefix probing evaluates a model's ability to generate similar continuations to a piece of text given the first $N$ tokens as context. In practice, to overcome the limitations of a model not generating an exact continuation, evaluation can be undertaken by measuring number of words in the longest common subsequence between the original text and the model generation (LCS) ~\cite{karamolegkou-etal-2023-copyright}. See Appendix E for examples of model-generated continuations and details about our prefix probing setup. For dataset contamination where samples are typically shorter and less amenable to prefix probing, we compare against TS-SLOT Guessing~\cite{deng2023investigating}, which probes contamination in black-box LLMs by first asking ChatGPT to identify ''informative'' words and then using these to make masked-prompts for the target model.

%% file: latex/sections/experiments_fiction.tex
\subsection{Fiction}
Memorization of fiction books has recently been studied due to the potential legal consequences of LLMs reproducing copyrighted texts~\cite{karamolegkou-etal-2023-copyright}. 
To test our probes, we examine the results of three types of surprise tokens: character names (\emph{Person}) which are known to be high-surprisal in fictional text~\cite{chang2023speak}, low probability tokens from a reference model (\emph{Prob}), and high rank tokens  (\emph{Rank}) from a reference model. For prefix probing, we use the first 50 words in the passage as input for the target model (Appendix E). 

 We use the BookMIA dataset~\cite{shi2023detecting}, which consists of text excerpts from books published in 2023 (after the knowledge cutoff of the models we study) as examples of unseen text, and passages from popular books as memorized examples~\cite{chang2023speak}. We use 8k examples from the dataset as a test set. We consider a sample of text to be memorized if at least two high-surprisal tokens are reconstructed successfully by the target model, to avoid the effect of a single spurious match (Appendix A).

\input{tables/token_examples}

\paragraph{Results.} 
Table \ref{tab:fiction} shows the precision, recall, and $F_{\beta=0.1}$ performance for Llama-2-70B, GPT-3.5, and GPT-4. 
Our goal is to minimize the number of false positive samples that are reported as memorized text.
Compared to suffix completions, we find that surprisal tokens can \emph{more precisely} identify memorized book passages for all three types of models. 
Further, we find that surprisal tokens based on domain knowledge (\emph{Person}) can be highly informative when available, though tokens obtained from reference models are also informative. 

We additionally show examples of the  high-surprisal tokens, selected either by rank or probability from, in Table \ref{table:example-tokens}.
These examples highlight the diversity of words selected, which span from rare tokens to character names to fictional place names to frequent tokens used in unusual settings.
These high-surprisal tokens would not easily be found via rule-based systems and can be applied across many different text domains. 
Even if no character name is available in a text, the surprisal metric can still be used to identify other tokens for the cloze test.

\input{tables/results_fiction_sample.tex}

%% file: tables/token_examples.tex

\begin{table}[t]
    \centering
    \small
    \begin{tabular}{@{}p{7.6cm}@{}}
        \toprule
        Tokens Selected by \textbf{Probability} 
        \\
        \midrule
        ...over Avathar, or Araman, or Valinor, and plunge in the chasm beyond the Outer Sea, pursuing his way alone amid the \underline{grots} and caverns at the roots of Arda... 
        \\[1ex] 
        Another sigh, and she let herself fall back, head cradled by the soft \underline{loam}. Her eyes closed...
        \\[1ex] 
        Anne breathed deeply, and looked into the clear sky beyond the dark \underline{boughs} of the firs.
        \\[1ex] 
        You’re safe. \underline{Benjamin} can’t hurt you anymore.
        \\[1ex]
        \toprule
        Tokens Selected by \textbf{Rank} 
        \\
        \midrule
        That light lives now in the \underline{Silmarils} alone. But Morgoth hated the new lights...
        \\[1ex]
        He slammed down another dollar. ``Don’t \underline{oversport} yourself, Ed,'' Bootyny challenged.
        \\[1ex]
        Because I am committed to protecting my peace and you are so far from my inner circle you’re basically a \underline{hexagon}. Get thee behind me.
        \\[1ex]
         Why, said Stubb, eyeing the velvet vest and the watch and seals, you may as well begin by telling him that he looks a sort of \underline{babyish} to me, though I don't pretend to be a judge.
        \\[1ex]
        \bottomrule
    \end{tabular}
    \caption{
    Examples of tokens detected as \textit{surprising} by rank and probability. These examples highlight the diversity of tokens and texts, which go beyond character names or other traditional cloze tests.
    }
    \label{table:example-tokens}
\end{table}

%% file: tables/results_fiction_sample.tex
\begin{table}[t]
\centering
\footnotesize
\begin{tabular}{llrrr}
    \toprule
    Probe & Token Type & P & R & $F_{\beta}$ \\ 
    \midrule
    Random & -  & 50.2  & 50.5  & 50.2  \\ 
    Majority  & -  & 49.8  & 100  & 50.2  \\ 
    \midrule
    \rowcolor[gray]{0.90} \multicolumn{5}{c}{\textit{Target Model}: GPT-3.5} \\
    LCS &- & 53.3 & 69.7 & 53.4  \\
    Surprisal & \emph{Person}  & 83.1 & 47.2 & \textbf{82.5 } \\ 
    Surprisal  & \emph{Prob} & 75.8  & 11.7  & \underline{71.9 }  \\ 
    Surprisal  & \emph{Rank} & 73.5 &10.5 & 69.3  \\ 
    \midrule
    \rowcolor[gray]{0.90} \multicolumn{5}{c}{\textit{Target Model}: GPT-4} \\ 
    LCS  &- & 56.8 & 63.6 & 56.9  \\
    Surprisal & \emph{Person} & 82.2 &75.3 & \underline{82.7 }\\ 
    Surprisal & \emph{Prob} & 81.9 & 61.8 & 81.6   \\ 
    Surprisal  & \emph{Rank} & 82.6 & 63.9 &\textbf{82.3}  \\ \midrule
    \rowcolor[gray]{0.90} \multicolumn{5}{c}{\textit{Target Model}: Llama-2-70B} \\ 
    LCS &- & 53.2 & 34.7 & 52.9  \\
    Surprisal  & \emph{Person} & 75.8 & 29.6 &  \textbf{74.6}\\ 
    Surprisal  & \emph{Prob} & 64.9 & 8.5 & \underline{60.9}  \\ 
    Surprisal  & \emph{Rank} & 64.4 &7.0 & 59.6  \\ 
    \bottomrule
\end{tabular}
\caption{Identification results for GPT-3.5 (top), GPT-4 (centre), and LLama-2-70B (bottom) on \textit{fictional} text, with $\beta$=0.1. We bold the \textbf{highest values} and underline the \underline{second highest}. }
\label{tab:fiction}
\end{table}

%% file: latex/sections/experiments_nyt.tex
\subsection{New York Times Lawsuit}

\input{tables/results_nyt}
In 2023, the New York Times sued OpenAI for allegedly training on articles published by the Times~\cite{nytlawsuit}. We scrape the evidence included in Exhibit-J of the New York Times lawsuit against OpenAI~\cite{nyt-exhibit-j}, consisting of one hundred articles that GPT-4 allegedly memorized. We also compare to negative samples, gathered from scraping hundred articles from CNN in 2023, which appear after the reported knowledge cutoff date for both models. For prefix probing, we use prefixes provided in Exhibit-J of the lawsuit (Appendix E).


\paragraph{Results.}
As shown in Table \ref{tab:open-domain-auto}, we evaluate both GPT-3.5 and GPT-4 on the resulting dataset, and use the Mistral-V2~\cite{jiang2023mistral7b} and Alpaca-7B~\cite{taori2023alpaca} models as knowledge filters. We find that (1) though the evidence in the lawsuit is based on near-exact verbatim regurgitation of the content of the New York Times articles --- this content is no longer exactly reproduced as of May 2024, likely because of additional post-training procedures or output filters. (2) We find that while verbatim prompting works better on GPT-3.5 with very few correct guesses on surprise tokens, probing with surprise tokens is much more effective when it comes to GPT-4 where the probe shows fewer false positives.

%% file: tables/results_nyt.tex
\begin{table}[t]
\centering
\footnotesize
\begin{tabular}{lrrrr}
    \toprule
    Probe & Token Type & P & R & $F_{\beta}$ \\ 
    \midrule
    Random & - &48.7 & 55.0 & 48.7 \\ 
    Majority & - & 50.0 & 100 & 50.3\\ 
    \midrule
    \rowcolor[gray]{0.90} \multicolumn{5}{c}{\textit{Target Model}: GPT-3.5} \\
    LCS & - & 59.8  & 61 & \textbf{59.82} \\
    Surprise Token & \textit{Rank} & 53.5 & 23 & \underline{52.8} \\
    Surprise Token & \textit{Prob}  & 51.5 & 17 & 50.5 \\
    Surprise Token+IF & \textit{Rank} & 37.5  & 3  & 33.7 \\
    Surprise Token+IF & \textit{Prob} & 40.0 & 4  & 36.7 \\
    \midrule
    \rowcolor[gray]{0.90} \multicolumn{5}{c}{\textit{Target Model}: GPT-4} \\
    LCS & - & 46.7 & 50 & 46.8 \\
    Surprise Token & \textit{Rank} & 64.7 & 44 & \underline{64.4} \\
    Surprise Token &  \textit{Prob} & 58.1 & 36 & 57.7 \\
    Surprise Token+IF & \textit{Rank} & 70.0 & 14 & \textbf{67.3} \\
    Surprise Token+IF & \textit{Prob} & 54.8 & 17 & 53.7 \\ 
    \bottomrule
\end{tabular}
\caption{Identification results for GPT-3.5 and GPT-4 on articles from the New York Times lawsuit, with $\beta$=0.1. IF indicates application of a knowledge filter.  We bold the \textbf{highest values} and underline the \underline{second highest}.}
\label{tab:open-domain-auto}
\end{table}

%% file: latex/sections/experiments_contamination.tex
\subsection{Dataset Contamination}

\input{tables/gpqa_contamination}
\input{tables/all_dataset_contamination_all}

A growing concern for accurate evaluation of LLMs' generalization capabilities is the prospect of dataset contamination, which is when evaluation benchmark data has already appeared in LLM training sets. 
Past work has demonstrated that this can artificially inflate benchmark performance~\cite{touvron2023llama2, zhou2023don, jiang2024investigating}. 
However, with proprietary models that limit access to training data, there is limited recourse to evaluate how contamination affects performance.

For our purposes, we study \emph{text contamination}~\cite{jiang2024investigating}, when the input text of an evaluation sample is likely to have appeared in the training data of a model.
There is little gold standard evidence of contamination for proprietary language models. Recently, \newcite{deng2023investigating} looked for evidence of contamination by prompting ChatGPT to identify informative words in dataset samples and prompting target models to guess the missing words (TS-SLOT). Specifically, they conduct a contamination experiment, where ChatGPT is finetuned with data from the MMLU test set, and the differences in Exact Match Rate between the finetuned model and the original model is observed.  However, the extent to which these target models were contaminated with those datasets in the first place is unknown, and consequently the discriminative power of the test also remains unknown. In the section below, we describe a controlled setting to examine the discriminative power of contamination tests. We then apply the best-performing tests to real-world datasets to surface evidence of contamination.


\paragraph{Controlled contamination.}  
We would like any test of contamination to have \emph{discriminative} power, i.e., produce different values for contaminated datasets and uncontaminated datasets.  

Therefore, we construct a synthetic test to compare the power of reconstruction probing with TS-SLOT~\cite{deng2023investigating}. 
We start with an uncontaminated dataset, and then deliberately contaminate the model to examine the discriminative power of our method at detecting contamination. Thus, we consider a benchmark dataset, Google-proof QA or GPQA~\cite{rein2023gpqa},  where questions were written by domain-experts in biology, physics, and chemistry. The dataset was released in 2023--- beyond the knowledge cutoff of the target models in this study.  The combination of novel written text and the release date of the benchmark\footnote{And not only the release date, as benchmarks may be based on text which predate the benchmark itself, such as Wikipedia articles from previous years.} leads to our decision to consider GPQA as unlikely to be contaminated for these models.

We then replicate the contamination experiment from ~\cite{deng2023investigating} as described in Table \ref{ref:gpqa-contamination-tsslot}, by finetuning GPT-3.5-turbo on the dataset. We use the Mistral-V2~\cite{jiang2023mistral7b} model as a knowledge filter. We observe that all surprisal-based approaches we study have greater discriminative power than TS-Slot-based approaches, with low-probability surprisal tokens with instruction-filtering having the greatest discriminative power.

\paragraph{Probing for contamination.}

We then apply this method on two other datasets (CommonsenseQA~\cite{talmor-etal-2019-commonsenseqa} and ARC-Challenge~\cite{allenai:arc}) to probe for evidence of contamination across our three target models (Table \ref{ref:all_contamination}). We apply these probes on the test sets for both datasets.

We find that we are largely unable to find evidence of contamination for the models we study on CommonsenseQA. For Arc-Challenge, we do observe slight evidence of contamination with GPT-4-0613. Qualitative analysis indicates that some examples of high-surprisal tokens are correctly predicted by GPT-4-0613, such as ``\emph{photosynthesis}'' in ``\emph{Which of these is produced during [MASK]?}'', and ``\emph{HCL}'' in ``\emph{If [MASK] is added to Zn, what would be an expected product?}''

%% file: tables/gpqa_contamination.tex



\begin{table*}[tb]
\resizebox{\textwidth}{!}{
\begin{tabular}{lrrrrr} 
\toprule
                            &   \makecell{TS-SLOT (EM)\\ ~\cite{deng2023investigating}} & \makecell{Reconstruction\\Probing \emph{Prob} (EM)} & \makecell{Reconstruction\\Probing  \emph{Rank} (EM)} & \makecell{Reconstruction\\Probing \emph{Prob IF} (EM)} & \makecell{Reconstruction\\Probing  \emph{Rank IF} (EM)} \\ 
\midrule
\makecell{\#Tokens}              &448  &          448                 & 448 & 258 & 207\\ 
\midrule
\makecell{gpt-3.5-turbo}              &     38.84\%          &     16.96\%                    & 12.28\% & 6.59\% & 7.25\% \\ 
\makecell{gpt-3.5-turbo\\(contaminated)} &    88.84\%           &  89.51\%                         & 66.07\% & 84.82\% & 78.64\%  \\ 
\midrule
\makecell{${\Delta}$} &     50\%       &  \underline{72.55\% }                        & 53.79\% & \textbf{78.23\%} & 71.39\% \\ 
\midrule
\end{tabular}
}
\caption{Exact Match rates for TS-SLOT~\cite{deng2023investigating} and information-guided probing for ChatGPT, with and without contamination. IF indicates application of a knowledge filter. The discriminative power of reconstruction probing is greater in all settings, including picking low probability tokens or tokens with a large number of alternatives. }
\label{ref:gpqa-contamination-tsslot}
\end{table*}

%% file: tables/all_dataset_contamination_all.tex



\begin{table}[]
\resizebox{\columnwidth}{!}{
\begin{tabular}{lrrr}  
\toprule
Dataset/Model        & Llama-2-70B & GPT-3.5 & GPT-4 \\ \midrule
\textcolor[gray]{0.6}{\# Tokens} & \textcolor[gray]{0.6}{104/86} & \textcolor[gray]{0.6}{104/86} & \textcolor[gray]{0.6}{104/86}\\
CommonsenseQA & 0.96\%/0.0\% & 6.73\%/2.33\% & 4.81\%/1.16\% \\ \midrule
\textcolor[gray]{0.6}{\# Tokens} & \textcolor[gray]{0.6}{105/51} & \textcolor[gray]{0.6}{105/50} & \textcolor[gray]{0.6}{105/51}\\
ARC & 11.43\%/1.96\% & 24.76\%/6.0\% & 22.86\%/13.73\%       \\ 

\bottomrule
\end{tabular}
}
\caption{Exact Match Rates for information-guided probing on two datasets: CommonsenseQA and ARC. We report \textcolor[gray]{0.6}{\# low-probability probes/\# of low-probability probes that pass instruction filtering}, as well as the exact match rates on these probes. }
\label{ref:all_contamination}
\end{table}

%% file: latex/sections/analysis.tex
\section{Analysis}

\paragraph{Memorization extraction methods are complementary.}

Overall, our experiments show that \emph{all three tested methods uncover distinct examples of memorized text that are not identified by any of the other methods in this study}.
This indicates that a suite of complementary methods may be most suitable to uncover the training data known to black-box large language models.

\paragraph{Larger models reconstruct more tokens.}
We examine the effect of model size on the capability to recover high-surprisal tokens (Figure 3). For the Llama-2 family of models, we plot the proportion of high-surprisal tokens recovered by the model at 7B, 13B and 70B parameters for samples from fictional text. We observe that larger models can recover many more high-surprisal tokens ($\sim$7x more tokens for the 70B model compared to the 7b model), indicating that surprisal probes are likely to be much more effective at recovering memorized data on large models compared to small models.

\paragraph{Token probabilities contain more information about training data.}
\input{tables/results_logprob.tex}

We would like to understand the upper bound on performance for identifying training data in LLMs. We therefore focus on a large model that does allow practitioners to access token probabilities on fictional text: Llama-2-70B (Table 6). Our work relates closely to membership inference methods, which seek to identify if data has been used to train a model, or if a model has never seen the data. Memorized training examples would form a subset of the model's training data. Thus, we study three kinds of membership inference methods: the perplexity of the sample, the perplexity callibrated with the zlib compression entropy~\cite{carlini2021extracting} and the Min-K method~\cite{shi2023detecting}. We find that the gap between methods that have full access to token probabilities, and surprisal-based methods is about 17 points in $F_{\beta}$. Notably, token probabilities represent a rich source of information in two aspects: (1) they are from the target model itself, (2) they are drawn from a much larger sample space than the binary information provided by whether a model could reconstruct a token (or not). When available for models, we advocate using token-probability based methods for membership inference.









\begin{figure}[tb]
    \centering
    \begin{subfigure}{0.3\textwidth}
        \centering
        \includegraphics[width=\linewidth]{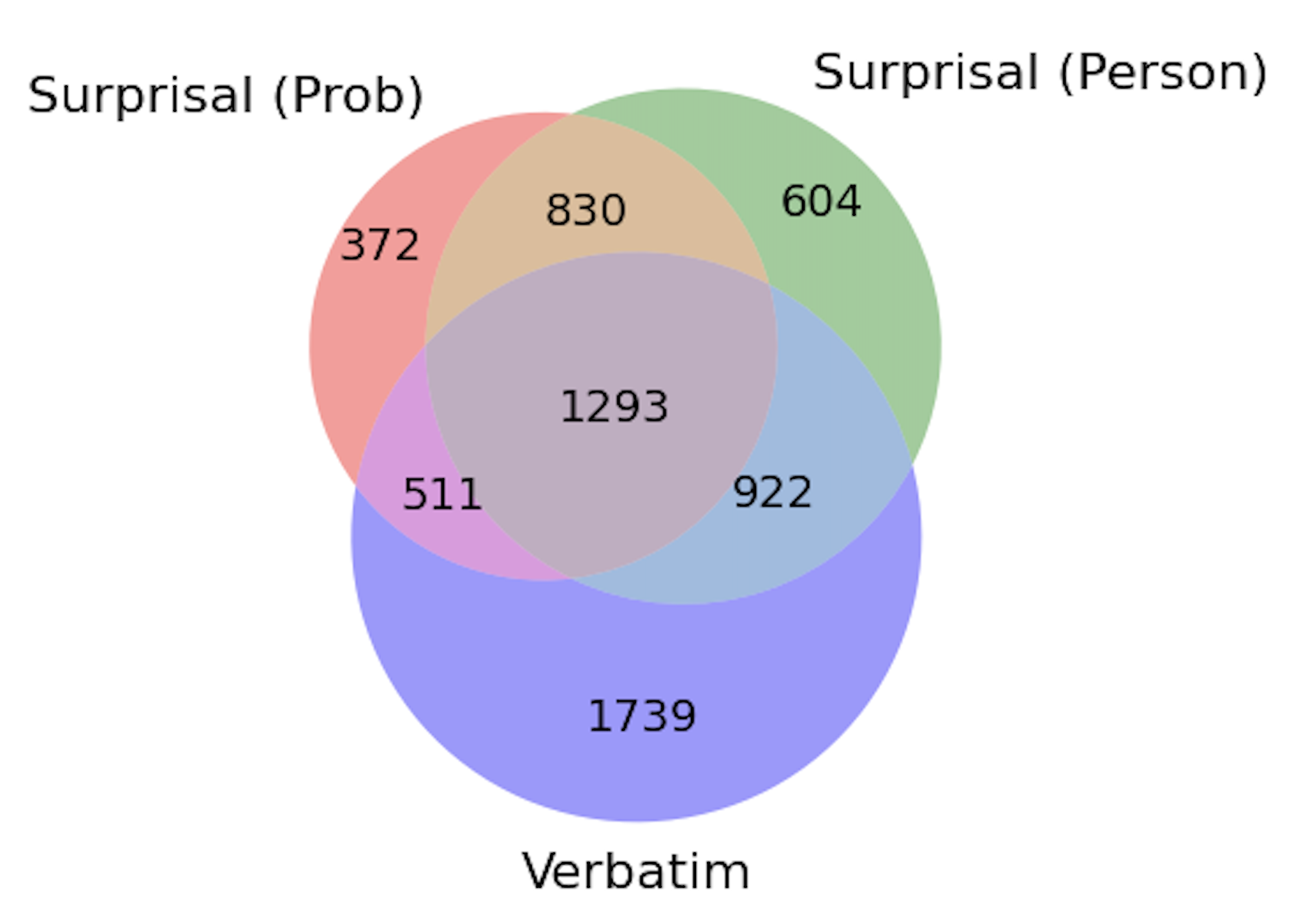}
        \caption{Overlap between all instances identified as memorized by GPT-4 on fictional text, for prefix probing (verbatim), and surprisal (person and prob)}
        \label{fig:1}
    \end{subfigure}
    \hfill
    \begin{subfigure}{0.3\textwidth}
        \centering
        \includegraphics[width=\linewidth]{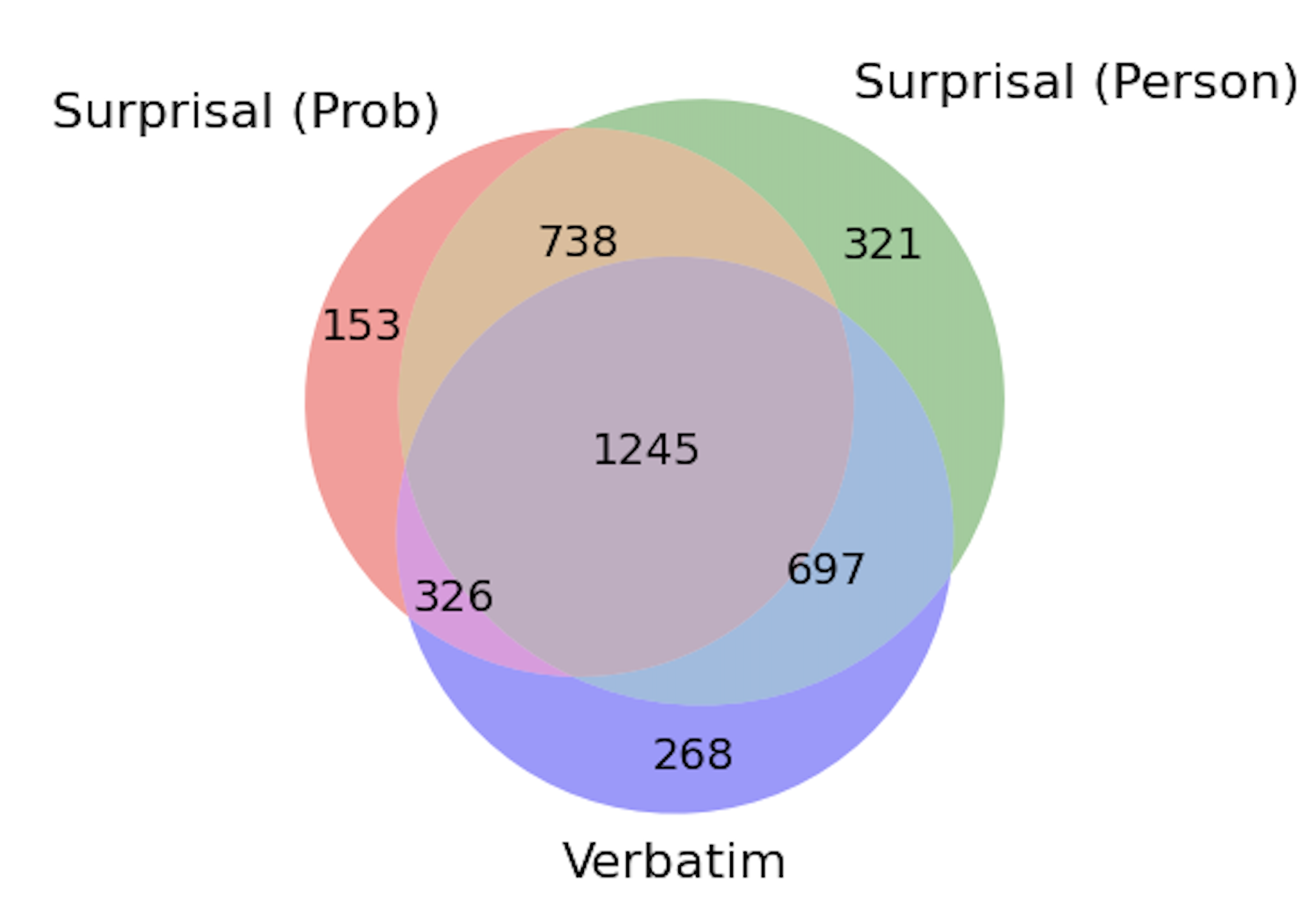}
        \caption{Overlap between all instances \emph{correctly} identified as memorized by GPT-4 on fictional text, for prefix probing (verbatim), and surprisal (person and prob)}
        \label{fig:2}
    \end{subfigure}
    \hfill
\caption{Overlap between instances identified as memorized for GPT-4, by surprisal-based probes and verbatim probes}
    \label{fig:overall_overlap}
\end{figure}

\begin{figure}[ht]
    \centering
    \includegraphics[width=\columnwidth]{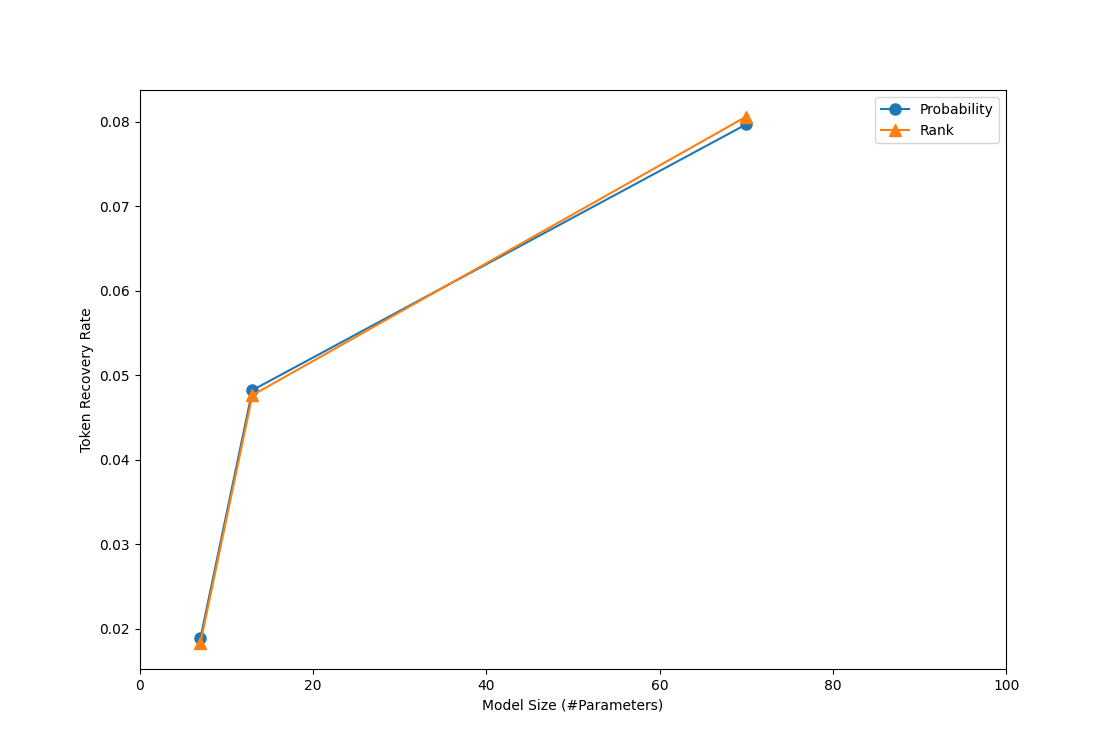}  
    \caption{Token recovery rate as a function of model size for Llama-2 7B, 13B, and 70B. We observe that as model size increases, the ability to recover low-likelihood tokens also increases.}
    \label{fig:sample_parametersize}
\end{figure}



%% file: tables/results_logprob.tex
\begin{table}[]
\resizebox{\columnwidth}{!}{
\begin{tabular}{lrrr}  
\toprule
Probe                                          & \multicolumn{1}{l}{Precision} & \multicolumn{1}{l}{Recall} & \multicolumn{1}{l}{F-Beta} \\ \midrule
\cellcolor[HTML]{FFFFFF}Surprise Tokens (Rank) & 64.4                         & 7.0                          & 59.6                      \\
\cellcolor[HTML]{FFFFFF}Surprise Tokens (Prob) & 64.9                         & 8.5                       & 60.9                       \\
Surprise Tokens (Person)                       & 75.8     & 29.6  & 74.6 \\ \midrule
PPL                                            & 96.5                          & 9.7                       & 88.6                      \\
PPL/ZLib                                       & 98.8                         & 9.9                       & 90.7                       \\
Min 5\%                                        & 99.5                          & 10.0                       & 91.4                      \\
Min 10\%                                       & 98.5                          & 9.9                       & 90.5                      \\
Min 20\%                                       & 98.0                            & 9.8                       & 90.0                      \\
Min 30\%                                       & 97.3                         & 9.8                       & 89.3                      \\
Min 40\%                                       & 96.8                         & 9.7                       & 88.9  \\ \bottomrule                   
\end{tabular}
}
\caption{Comparison of information-guided probing to membership inference methods that access token probabilities, for Llama-2-70B. Membership inference methods can be viewed as an upper bound --- both seeking to identify all training data and non-members (and not just memorized training data), and having access to the token probability distribution from the target model. }
\end{table}

%% file: latex/sections/discussion.tex
\section{Discussion and Future Work}
 Our goal is to provide a foundation for greater data transparency in the ecosystem surrounding large language models. We briefly discuss our findings, and provide directions for future research. 

\paragraph{The need for multiple methods to surface training data.} In our work, we find that we are able to identify text that is known to even proprietary black-box LLMs, and that the examples of memorized text that were successfully identified can differ between probing methods. This indicates that the community would benefit from a range of such approaches, and that focusing on state-of-the-art detection performance should not be the only goal. Further, recent work has investigating combining signals for various training data identification methods in order to determine if a model was trained on a given document~\cite{maini2024llm}. This suggests that developing diverse, complementary, probes can help us better understand how data was used to train models.

\paragraph{Answering questions about model generalization.} While LLMs have achieved state-of-the-art performance on several benchmarks, the extent of their generalization capabilities remains an open question. This is in part due to an inability to characterize train-test overlap: that is, making sense of what data a model was trained on and how it may relate to the data a model needs to perform inference on. Our work adds to the growing body of work in shedding light on test-set contamination, which is particularly critical for proprietary LLMs --- where the reasons behind performance improvement on standard benchmarks are opaque.

\paragraph{Data transparency in the LLM ecosystem.} Our work is in spirit related to studies that  have called for better data documentation in machine learning ~\cite{liu2024automatic,bender-friedman-2018-data,gebru2021datasheets} or offered large-scale search and indexing mechanisms for pre-training corpora~\cite{elazar2023s,liu2024infini}. While previous efforts focus on supervised, accessible datasets, this work focuses on inferring training data of models, especially in cases where there is currently no data transparency.

\paragraph{Why not only work with open models?} In this work, our central focus is closed commercial models that do not offer access to token probabilities. The reason for this choice is that this is the dominant paradigm for many popular models today. While there is an argument to be made to limit scientific study only to open-source models, studying proprietary models is also important for multiple reasons: (1) they are in frequent use by the general public, and it is critical to understand how they may be using people's data, (2) open models are often required to match the capabilities of proprietary models in order to be considered in the same class, so it is important to understand where performance improvements of proprietary models comes from.

\paragraph{Influence of data on models.} Little is known about how datapoints in training affect model behavior downstream. A systematic understanding of these relationships could enable more principled approaches to data curation, and help optimize for specific behaviors by identifying which types of training examples are most effective for it.  We hope by identifying training data that is memorized by models, we enable further studies of how data affects model behavior, and which data most affects model behavior.









%% file: latex/sections/related_work.tex
\section{Related Work}

Our work closely relates prior lines of work in the literature that involve training data exposure of models, and data transparency. Typically, such work has either examined copyright issues, or investigated evidence of dataset contamination--- our study explores both of these issues, and we discuss prior work in both here.


\paragraph{Copyright violation risks.}
Copyright issues can arise at various steps in the generative AI pipeline, especially in language models, including data collection~\cite{min2023silo,shi2023detecting,chang2023speak,karamolegkou-etal-2023-copyright}, model training~\cite{vyas2023provable}, and generation and deployment~\cite{meeus2024copyright,ippolito2022preventing}. Our work relates to the first step by surfacing evidence of training data memorization from the models output, with only API-level access. ~\citet{duarte2024cop} propose DE-COP, a membership inference method that can work on black-box LLMs. This method is intended for document-level membership inference which is not the focus of our study, and works by aggregating evidence across passages in a longer document. This method is also expensive and relies on proprietary LLMs (see Appendix C,D for cost and performance comparisons). 

\paragraph{Dataset contamination.}
Work involving data contamination and test-case leakage have garnered more attention recently as such contamination could muddy the conclusions made from existing benchmarks~\cite{oren2023proving,golchin2023time,weller2023according,xu2024benchmark,sainz-etal-2024-data}.
Although this line of work also infers membership, it differs from our work in two manners: (1) the information-guided probes can be applied to even proprietary large language models with no access to token probabilities from the model, and (2) test-set contamination methods sometimes take advantage of meta-data and artifacts other than the data itself, for instance the order of samples~\cite{oren2023proving}, whereas in our mode we do not have access to such meta-data.

\paragraph{Verbatim memorization and membership inference.}

Our work is also related to membership inference attacks (MIAs), which
 are often used as a proxy to measure the amount of training data leakage in machine learning models~\citep{shokri2017membership}.
These attacks usually entail thresholding a membership score, which is metrics
 including LOSS~\cite{yeom2018privacy}, likelihood-ratio ~\cite{carlini2021extracting,mireshghallah2022quantifying}, Zlib Entropy~\cite{carlini2021extracting}, curvature~\cite{mattern2023membership}, and Min-k\% probability~\cite{shi2023detecting}, among others. More recent work~\cite{duan2024membership} has shown that membership inference attacks for LLMs show near-random performance, partly due to models being trained on large datasets with very few iterations.  In contrast, our work seeks to find evidence of memorization of datapoints, and identify data that has left a strong imprint in the model. Previously, evidence of such memorization has largely been found by examining model generations for long sequences that are likely from the training data~\cite{carlini2021extracting}, or by prompting the model to generate continuations of a piece of text given the first part of text as input~\cite{karamolegkou2023copyright}. Modern LLMs likely incorporate additional posttraining or output filters to safeguard against verbatim regurgitation, indicating the need for a suite of complementary methods to uncover evidence of memorization, this present work offers a unique and complementary approach to identify this evidence.
 



%% file: latex/sections/appendix.tex
\section{Knowledge Cutoffs and Hyperparameters}
We select all target models for this study (\texttt{gpt-4-0613}, \texttt{gpt-3.5-turbo-0125}, and Llama-2-70B) such that they have a knowledge cutoff date before 2023.

We use a validation set of 1870 examples from BookMIA to be used for tuning all hyperparameters, which is held out separately from the 8000 examples we evaluate on. We select surprisal tokens which have a log likelihood $<$-12, or a rank$>$2000. The number of high-surprisal tokens can vary in any given piece of text, we cutoff at 10 tokens for text where more than 10 high-surprisal tokens are available (selecting the lowest probability or highest rank tokens that pass the threshold). For fiction datasets, we consider a sample to be memorized if atleast two or more such tokens can be reconstructed by the target model (to overcome the effect of a single spurious match). For the New York Times dataset, we additionally use knowledge filters since it is a real-world domain.







\section{Prompts}

Inspired by ~\cite{chang2023speak}, we use the following prompt for probing fictional text and NYT articles:

\begin{tcolorbox}[colback=gray!10, colframe=gray!50, sharp corners, fonttitle=\scriptsize, breakable]
\scriptsize 
You have seen the following passage in your training data. What is the word that fills in the [MASK] token in it? This word is a noun, verb, adjective, or adverb (not a  pronoun, preposition or conjunction). You must make a guess, even if you are uncertain.\\

Example:

Input: Stay gold, [MASK], stay gold.

Output: <word>Ponyboy</word>\\

Input: The door opened, and [MASK], dressed and hatted, entered with a cup of tea.

Output: <word>Gerty</word>\\

Input:
\end{tcolorbox}

Inspired by ~\cite{deng2023investigating}, we use the following prompt for test set contamination:

\begin{tcolorbox}[colback=gray!10, colframe=gray!50, sharp corners, fonttitle=\scriptsize, breakable]
\scriptsize 
Complete the sentence in one word:

Input:

Reply the answer only.
\end{tcolorbox}

We use the following two prompta for knowledge filtering.

\begin{tcolorbox}[colback=gray!10, colframe=gray!50, sharp corners, fonttitle=\scriptsize, breakable]
\scriptsize 
What is the word that fills in the [MASK] token in the following passage?

Passage:
\end{tcolorbox}

\begin{tcolorbox}[colback=gray!10, colframe=gray!50, sharp corners, fonttitle=\scriptsize, breakable]
\scriptsize 
What are 100 words that can fill in the [MASK] token in the following passage?

Passage: 
\end{tcolorbox}

\section{Probe Cost}

We attempt to estimate the cost of running information-guided probes. For a given input of $N$ tokens, this cost would scale with the number of high-surprisal tokens $K$ being probed. Let us assume that these tokens are cutoff at $K=10$ as we do in this work, and let us assume output tokens are constrained to 5 tokens (though models may provide longer responses). Then for any provider where input tokens cost X\$/token, and output tokens cost Y\$/token (for many LLM providers X$<$Y), the cost of running information-guided probes would be $10*((N+123)*X + 5*Y)$, where 123 is the number of tokens in the instructions for probing fictional text using the tokenizer for GPT-4. In comparison, prefix probing~\cite{karamolegkou-etal-2023-copyright} would cost $(50+13)*X+(N-50)*Y$, assuming that 50 tokens are provided to the model as a prefix. We estimate DE-COP~\cite{duarte2024cop} for a sample would cost $N*4*24*X + 24*Y$, not taking into account the cost of paraphrasing the input three times with Claude, making it more expensive than the other probes if $N>15$.
\section{Adapting DE-COP to Passages}

DE-COP~\cite{duarte2024cop} is a method intended to determine whether a piece of text has appeared in a model's training data. In the fully black-box setting, detecting copyrighted content is posed as a multiple-choice question answering task (MCQA) task, where a target model must correctly pick out the training sample, from three paraphrases of that training sample. Though this method is intended for document-level membership inference by aggregating information across several passages in a long document, we attempt to do a head-to-head comparison to information-guided probing by considering the DE-COP score assigned to each passage. We subsample 600 passages from the BookMIA dataset due to the cost of the DE-COP method, and then follow the original procedure of using Claude-2~\cite{claude2} to paraphrase each passage three times. The prompt used for paraphrasing is `Rewrite this entire text (all sentences with no exception) expressing the same meaning using different words. Aim to keep the rewriting similar in length to the original text. Do it three times. The text to be rewritten is identified as <Example A>. Format your output as: Example B: <insert paraphrase B> Example C: <insert paraphrase C> Example D: <insert paraphrase D> Example A:'. For each of these samples, we then oversample 24 times by creating every possible combination of options for the 4-option multiple choice question. To evaluate memorization, we compute the proportion of times the model picks the BookMIA passage instead of its paraphrased counterparts in the multiple-choice question answering task. We prompt the model with the following prompt that doesn't use any extra metadata about the sample `You are entering a multiple choice questions exam. You should directly answer each question by choosing the correct option. Be concise and straight to the point in your answer. Output only the letter corresponding to the correct answer.
Question: Which of the following passages is verbatim from your training data? Options:'. We use 100 examples from the same validation set to compute the threshold to identify memorized samples in DE-COP, and evaluate on the remaining 600 samples. We also evaluate information-guided probing on the same 600 samples. These results are shown in Table. \ref{tab:decop} for \textsc{GPT-3.5}. We find that in the same head-to-head setting where both probes only have access to a text snippet, information-guided probing is more precise.

\begin{table}[t]
\centering

\resizebox{1\columnwidth}{!}{
\begin{tabular}{lrrr}
    \toprule
    Probe  & Precision & Recall & $F_{\beta}$ \\ 
    \midrule
    Majority   & 51.5  & 100  & 51.75  \\ 
    DE-COP-Passage  & 58.38 & 63.11 & 58.43  \\
    Surprisal \emph{(Person)}  & 85.47 & 47.57 & \textbf{84.8 } \\

    \bottomrule
\end{tabular}
}
\caption{Identification results for GPT-3.5 (top) on 600 samples from the BookMIA dataset, with information-guided probing and DE-COP, with $\beta$=0.1. }
\label{tab:decop}
\end{table}



\section{Verbatim Completion and Prefix Probing}

\paragraph{Prefix Probing}
We describe the details of performing classification based on the prefix-probing approach wherein models are prompted with the first $N$ tokens of a piece of text and asked to continue it, and the generated continuation is then compared to the original piece of text. For fiction, we use the first 50 words as the prefix that is fed into the model, and for New York Times, we use the prefix specified in Exhibit-J of the New York Times vs OpenAI lawsuit for each article~\cite{nyt-exhibit-j}. For prefix probing with fiction, we add the instruction `Please complete the following text with more than 150 words:' before the prefix, based on the prompt used in \citet{karamolegkou-etal-2023-copyright}. For prefix probing on New York Times articles, we add the instruction 'Complete exactly and don't hallucinate' before the prefix, based on successful reproduction of parts of a New York Times Article using this prompt.\footnote{\url{https://x.com/srush_nlp/status/1740746983512609088}} We use a logistic regression classifier fit on the LCS values from the validation set to predict memorized samples on the test set. 

\paragraph{Examples of verbatim completions} We provide some qualitative examples of model completions to prefix probes in Table 8. We observe that models rarely generate the exact continuations as the original content in response to prefix probes, demonstrating the need for better metrics to assess model generations for evidence of memorization.
\input{tables/verbatim_completions}

\section{Books known to ChatGPT}
We include examples of high-surprisal tokens recovered from popular books by ChatGPT in Table 9. We find several such instances where tokens that are difficult to predict based on context alone without prior knowledge of the material, are successfully recovered by GPT-3.5 and GPT-4.
 \input{tables/book_title_tables}
\section{TS-SLOT}
We apply the filtering heuristics upon TS-SLOT, based on only keeping informative words that are nouns, adjectives, or verbs, and obtain only slightly different results (Table 10).
\input{tables/gpqa_contamination_full}

\section{Additional Surprisal Measures}

In this work, we consider two information measures for identifying high-surprisal tokens: probability, and rank. For fictional text, we additionally consider named entities. However, future work could find `surprising' tokens in multiple ways including (1) leveraging domain-specific knowledge to identify statistically rare or distinctive phrases (e.g., URLs, or usernames in an online forum--- character names are an example of this kind of heuristic.). (2) Leveraging n-gram statistics from large text corpora, (3) Manual annotations. In addition, future work could extend our method to identify surprising spans of text, or surprising structure in text to evaluate evidence of memorization.


%% file: tables/verbatim_completions.tex
\begin{table*}[!h]
\centering
\small
\begin{tabular}{p{0.08\textwidth}p{0.20\textwidth}p{0.3\textwidth}p{0.05\textwidth}p{0.25\textwidth}} \toprule
\textbf{Category}     & \textbf{Prompt} & \textbf{Original Continuation} & \textbf{Model} & \textbf{Model Response} \\ 
\toprule
Fiction & Please complete the following text with more than 150 words: Dexter said . It was hard to tell whether he actually believed what he was saying . “ Anything to rejuvenate the brand in Urmau , ” Brianna said , finally getting it . She stuffed a too-big forkful of salad in her mouth . Genial Sr. took a sip & of his soup and Manuel followed suit . Genial watched them nervously , scratching his cheek with his empty hand . “ Delicious , ” Genial Sr. pronounced , smiling at his son . “ You have a better chef than I do. ” “ Only the finest , ” Genial said , and looked at me as though I were the chef myself . “ It ’ s right , Ezra ? ” “ Only the finest , ” I echoed . The conversation stalled as we ate and were poured more wine . I entertained the thought that this dinner would last forever , that I would grow old and die at Genial ’ s giant dining room table  & \textsc{GPT-4}& of his aged whiskey, his eyes never leaving Dexter. The tension in the room was palpable, the air thick with anticipation. "Dexter," Genial Sr. began, his voice a gravelly whisper that commanded attention, "I've seen many a brand rise and fall in my time. It's not just about rejuvenation, it's about understanding the pulse of the market, the desires of the consumers." ...  \\ \midrule
Fiction & Please complete the following text with more than 150 words: not certain whether his sister had been born then . Finally they had emerged into a noisy , crowded place which he had realized to be a Tube station . There were people sitting all over the stone-flagged floor , and other people , packed tightly together , were sitting & on metal bunks , one above the other . Winston and his mother and father found themselves a place on the floor , and near them an old man and an old woman were sitting side by side on a bunk . The old man had on a decent dark suit and a black cloth cap pushed back from very white hair : his face was scarlet and his eyes were blue and full of tears . He reeked of gin . It seemed to breathe out of his skin in place of sweat , and one could have fancied that the tears welling from his eyes were pure gin . But though slightly drunk he was also suffering under some grief that was genuine and unbearable . In his childish way Winston grasped that some terrible thing , something that was beyond forgiveness and could never be remedied , had just happened . It also seemed to him that he knew what it was . Someone whom the old man loved -- a little granddaughter , perhaps -- had been killed ...& \textsc{GPT-3.5}  & on the benches lining the walls. The air was thick with the smell of sweat and the sound of chatter and footsteps echoed off the tiled walls. The boy felt overwhelmed by the sheer number of people surrounding him, each lost in their own world, yet all connected by the shared experience of navigating the bustling underground network.As he stood there, trying to make sense of his surroundings, a wave of uncertainty washed over him. He couldn't remember how he had ended up in the Tube station or where he was supposed to be going. Panic started to rise within him, but he forced himself to stay calm and think logically. He scanned the station \\ 
 \\ \midrule
New York Times & Complete exactly and don't hallucinate. WASHINGTON — While senior Trump administration officials said this week that Iran has been actively interfering in the presidential election, many intelligence officials said they remained far more concerned about Russia, which in recent days has hacked into state and local computer & networks in breaches that could allow Moscow broader access to American voting infrastructure. The discovery of the hacks came as American intelligence agencies, infiltrating Russian networks themselves, have pieced together details of what they believe are Russia’s plans to interfere in the presidential race in its final days or immediately after the election on Nov. 3. Officials did not make clear what Russia planned to do, but they said its operations would be intended to help President Trump, potentially by exacerbating disputes around the results, especially if the race is too close to call... & \textsc{GPT-4} & networks in an escalation of its sophisticated cyberattacks. The officials, who spoke on the condition of anonymity to discuss classified information, said that Russia's hacking efforts were aimed at disrupting the election process, sowing discord, and undermining confidence in the democratic process. They also noted that Russia has a history of interfering in U.S. elections, most notably in 2016. The Trump administration officials, on the other hand, have been focusing on Iran's alleged interference, accusing the country of sending threatening emails to voters and spreading disinformation online... \\ 
 \bottomrule    
\end{tabular}
\caption{Examples of verbatim completions by GPT-3.5 and GPT-4 to prefix probes, where part of the input is provided to the model and the model is asked to generate a continuation. We observe that we rarely see model responses that reflect the exact continuations. }
\label{ref:qreasoning_types}
\end{table*}

%% file: tables/book_title_tables.tex
\begin{table*}[!h]
\centering
\small
\begin{tabular}{p{0.1\textwidth}p{0.5\textwidth}p{0.1\textwidth}p{0.1\textwidth}} \toprule
\textbf{Book Title}     & \textbf{Probes} & \textbf{Masked Token} & \textbf{\textsc{GPT}-3.5 Response}\\ 
\toprule

\textbf{The Silmarillion} & n the white ships of the Teleri and landed in the Firth of Drengist, and there burned the ships at Losgar. Chapter 11 Of the Sun and Moon and the Hiding of Valinor It is told that after the flight of \textbf{[MASK]} the Valar sat long unmoved upon their thrones in the Ring of Doom; but they were not idle, as Fanor declared in the folly of his heart. For the Valar may work many things with thought rather than with hands, and without voices in sil & Melkor & Melkor \\ \midrule
\textbf{Emma} & a common way, like every body else, and if there is any thing to say, to sit down and write a letter, and say just what you must, in a short way; and another, to write verses and charades like this." \textbf{[MASK]} could not have desired a more spirited rejection of Mr. Martin's prose. "Such sweet lines!" continued Harriet--"these two last!--But how shall I ever be able to return the paper, or say I have found it out?--Oh! Miss Woodhouse, what ca & Emma & Emma \\ \midrule
\textbf{Great Expectations} & and disappeared. "Now, Mr. Pip," said Mr. Jaggers, "attend, if you please. You have been drawing pretty freely here; your name occurs pretty often in \textbf{[MASK]}'s cash-book; but you are in debt, of course?" "I am afraid I must say yes, sir." "You know you must say yes; don't you?" said Mr. Jaggers. "Yes, sir." "I don't ask you what you owe, because you don't know; and if you did know, you w & Wemmick & Output: Wemmick \\ 
\midrule 
\textbf{Hitchhiker's Guide To The Galaxy} &  "Yeah." "Er, what is ..." "A teaser? Teasers are usually rich kids with nothing to do. They cruise around looking for planets which haven't made interstellar contact yet and buzz them." "Buzz them?" \textbf{[MASK]} began to feel that Ford was enjoying making life difficult for him. "Yeah", said Ford, "they buzz them. They find some isolated spot with very few people around, then land right by some poor soul whom no one's ever going to believe a & Arthur & I believe the word that fills in the [MASK] token in the passage is "Arthur."\\
 \bottomrule    
\end{tabular}
\caption{Examples of book passages where GPT-3.5 recovered high-surprisal tokens}
\label{ref:book_title_types}
\end{table*}

%% file: tables/gpqa_contamination_full.tex



\begin{table*}[tb]
\resizebox{\textwidth}{!}{
\begin{tabular}{lrrrrr} 
\toprule
                            &   \makecell{TS-SLOT - Filtered (EM)\\ ~\cite{deng2023investigating}} & \makecell{Reconstruction\\Probing \emph{Prob} (EM)} & \makecell{Reconstruction\\Probing  \emph{Rank} (EM)} & \makecell{Reconstruction\\Probing \emph{Prob IF} (EM)} & \makecell{Reconstruction\\Probing  \emph{Rank IF} (EM)} \\ 
\midrule
\makecell{\#Tokens}              &442  &          448                 & 448 & 258 & 207\\ 
\midrule
\makecell{gpt-3.5-turbo}              &     38.91\%          &     16.96\%                    & 12.28\% & 6.59\% & 7.25\% \\ 
\makecell{gpt-3.5-turbo\\(contaminated)} &    88.91\%           &  89.51\%                         & 66.07\% & 84.82\% & 78.64\%  \\ 
\midrule
\makecell{${\Delta}$} &     50\%       &  \underline{72.55\% }                        & 53.79\% & \textbf{78.23\%} & 71.39\% \\ 
\midrule
\end{tabular}
}
\caption{We apply the additional filtering strategy proposed by ~\cite{deng2023investigating}, wherein only nouns, adjectives and verbs are retained (TS-SLOT-Filtered). We find this produces similar results on GPQA.}
\end{table*}